\documentclass{article}
\usepackage{ijcai24}

\usepackage{times}
\usepackage{soul}
\usepackage{url}
\usepackage[hidelinks]{hyperref}
\usepackage[utf8]{inputenc}
\usepackage[small]{caption}
\usepackage{graphicx}
\usepackage{amsmath}
\usepackage{amsthm}
\usepackage{booktabs}
\usepackage{algorithm}
\usepackage{algorithmic}
\usepackage[switch]{lineno}

\usepackage{algorithm}
\usepackage{algorithmic}
\usepackage{subfig}
\usepackage{multirow}
\usepackage{graphicx}
\usepackage[inkscapelatex=false]{svg}
\usepackage{pdfpages}
\usepackage{amsthm,amsmath,amssymb}
\usepackage{mathrsfs}

\title{Federated Knowledge Transfer Fine-tuning Large Server Model with Resource-Constrained IoT Clients}

\author{
Shaoyuan Chen$^{1,2}$
\and
Linlin You$^1$\thanks{Corresponding author: youllin@mail.sysu.edu.cn}
\and
Rui Liu$^3$
\and
Shuo Yu$^4$
\And
Ahmed M. Abdelmoniem$^5$
\affiliations
$^1$School of Intelligent Systems Engineering, Sun Yat-Sen University, Shenzhen, China\\
$^2$Shenzhen Fangle Technology Co., Ltd., Shenzhen, China\\
$^3$School of Computer Science and Engineering, Nanyang Technological University, Singapore\\
$^4$School of Computer Science and Technology, Dalian University of Technology, Dalian, China\\
$^5$School of Electronic Engineering and Computer Science,
Queen Mary University of London, UK\\
}

\begin{document}

\maketitle

\begin{abstract}
    The training of large models, involving fine-tuning, faces the scarcity of high-quality data. Compared to the solutions based on centralized data centers, updating large models in the Internet of Things (IoT) faces challenges in coordinating knowledge from distributed clients by using their private and heterogeneous data. To tackle such a challenge, we propose KOALA (Federated \textbf{K}n\textbf{o}wledge Tr\textbf{a}nsfer Fine-tuning \textbf{La}rge Server Model with Resource-Constrained IoT Clients) to impel the training of large models in IoT. Since the resources obtained by IoT clients are limited and restricted, it is infeasible to locally execute large models and also update them in a privacy-preserving manner. Therefore, we leverage federated learning and knowledge distillation to update large models through collaboration with their small models, which can run locally at IoT clients to process their private data separately and enable large-small model knowledge transfer through iterative learning between the server and clients. Moreover, to support clients with similar or different computing capacities, KOALA is designed with two kinds of large-small model joint learning modes, namely to be homogeneous or heterogeneous. Experimental results demonstrate that compared to the conventional approach, our method can not only achieve similar training performance but also significantly reduce the need for local storage and computing power resources.
\end{abstract}

\section{Introduction}

Models with ever-growing scale have been introduced, such as BERT ~\cite{b1:devlin2018bert,b2:liu2019roberta}, GPT ~\cite{b3:radford2018improving,b4:radford2019language,b5:brown2020language}, VGG ~\cite{b6:simonyan2014very}, and ViT ~\cite{b7:dosovitskiy2020image}. To train and adopt them in various Internet of Things scenes, how to utilize distributed data and computing powers becomes crucial. Unfortunately, IoT clients typically exhibit data protection considerations ~\cite{b8:chen2023federated,b9:zhuang2023foundation} and constrained computing capacities ~\cite{b10:wang2019edge,b11:imteaj2021survey}. These factors impede the use of their data to train complex and large-scale models. 

To tackle the challenge of data privacy, solutions based on federated learning (FL) are studied to support the training of large models in a collaborative and privacy-preserving manner, e.g., Yu S et al. ~\cite{b12:yu2023federated} propose a method of training the large model alternately in clients with private data and the server with labeled public data; and Wu C et al. ~\cite{b13:wu2022communication} introduce a method of federated mutual distillation for training personalized large models, which can significantly reduce communication costs. Even though private knowledge can be shared among distributed clients through FL, the common premise of current methods is to have sufficient local computing capacities to run large models directly on each learning client, making them infeasible to support distributed IoT clients with insufficient local resources.

\begin{figure}
    \centering
    \includegraphics[scale=0.52]{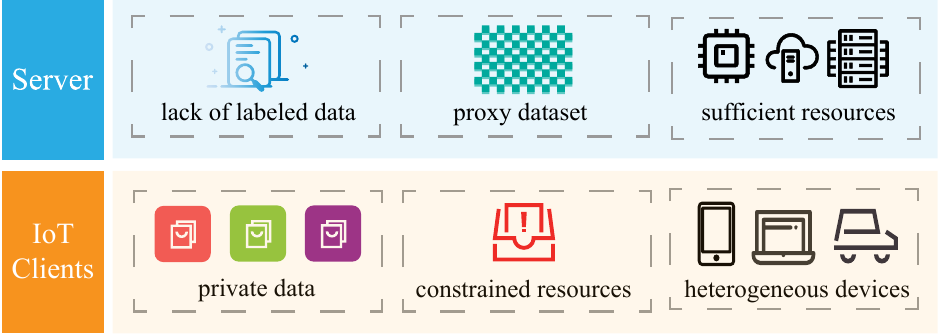}
    \caption{The situations of the server and IoT clients.}
    \label{fig:problems}
\end{figure}

Therefore, to support the fine-tuning of large models ~\cite{b14:houlsby2019parameter,b15:han2024parameter} and the model adaptation to empower various IoT scenarios, the objective of this study is defined as illustrated in Fig \ref{fig:problems}, where 1) the server has sufficient storage and computing powers but lacks high-quality data (only with a limited amount of unlabeled proxy dataset), and 2) IoT clients as a group are rich in sensed data and distributed computing powers, but as for each client, its device and private data are heterogeneous, and its local resources are limited to support the running of large models.

By integrating FL to share private knowledge across IoT clients and knowledge distillation (KD) to transfer encoded knowledge among different models (i.e., between teacher and student models), KOALA is proposed to enable a joint and iterative learning process that allows the IoT clients to run their local small models to extract and share local knowledge, and then the server to update the adapter of the large model based on the local updated small model of each client. Specifically, to implement such a learning process, the forward and reverse distillation techniques are used jointly, to, first, reverse distill trained small models to fine-tune the large model, and then, forward distill the large model to update small models for IoT clients. 

Moreover, in conventional FL, the global and local models have the same structure, and the global model can be updated based on the aggregation of local updates directly. However, the large-small model collaborative learning process implemented in KOALA needs to support different models in the server and clients, which makes conventional FL methods infeasible. Hence, according to the difference among small models, KOALA implements two kinds of learning mode to aggregate local knowledge encoded in homogeneous or heterogeneous small models. Specifically, the homogeneous method supports IoT clients to run small models with the same structure, and on the contrary, the heterogeneous method supports each IoT client to run different small models, which are more flexible as they can be created according to the actual computing capacity of the client. After the update of the large model, by using either homogeneous or heterogeneous methods, related small models can distilled from the latest large model and dispatched to their corresponding clients to start a new learning iteration. 

Based on standard datasets, the efficiency and effectiveness of KOALA are evaluated. Experimental results show that compared with the baseline, where IoT clients can load and execute the large model with sufficient local resources, our method can approach similar training performance for all tasks, and also significantly reduce the need for local resources. 

In general, our main contributions can be summarized as follows:
\begin{itemize}
\item[$\bullet$] We propose a novel large-small model collaborative learning process in data protection and resource-constrained IoT scenarios, through which, FL and KD can work jointly to support the iterative learning of large and small models even though they are cross-scale in model structures;
\item[$\bullet$] We design a reverse knowledge distillation strategy to better handle the outputs of heterogeneous small models updated based on local data, through which, the outputs of local models on proxy datasets are refined and integrated to generate consensus soft labels for large model fine-tuning;
\item[$\bullet$] The proposed method KOALA is verified to be performance-equivalent and resource-efficient. Specifically, large models fine-tuned by KOALA can achieve similar accuracy to the ones updated in conventional methods. At the same time, compared to conventional methods, the storage space needed for loading the local model reduces by about 97.6$\%$ (Homo) and 97.2$\%$ (Hete), and FLOPs of the local model reduces by about 98.4$\%$ (Homo) and 98.6$\%$ (Hete).  
\end{itemize}

\section{Related Work}

\subsection{Federated Learning}

Federated learning is a privacy-preserving machine learning framework where the server coordinates multiple clients to learn globally shareable models without exchanging local data directly ~\cite{b16:zhang2021survey}. As the classic method, FedAvg ~\cite{b17:mcmahan2017communication} manages each client to train its local model and upload the updated local model to the server. Then, the local models are aggregated to update a global model, which is then downloaded by active clients in the next round. However, the issue of non-identically and independently distributed (Non-IID) data among clients degrades the performance of federated learning ~\cite{b18:mora2022federated}, prompting numerous methods that aim to alleviate this problem. Accordingly, FedProx ~\cite{b19:li2020federated} introduces a proximal term to the loss function in local training, to constrain the updating of model parameters. SCAFFOLD ~\cite{b20:karimireddy2020scaffold} introduces control variables to reduce ``client drift''. MOON ~\cite{b21:li2021model} combines federated learning and contrastive learning to make the local model updating closer to the global model and farther away from the previous local model. Since highly heterogeneous data may prevent the model from converging, and a common global model fails to meet the individual needs of different clients, personalized federated learning is essential ~\cite{b22:tan2022towards}. FedClassAvg ~\cite{b23:jang2022fedclassavg} conducts federated learning on heterogeneous models through classifier aggregation. Per-FedAvg ~\cite{b24:fallah2020personalized} incorporates the classic meta-learning framework, MAML ~\cite{b25:finn2017model}, to train personalized models based on the global meta-model. Differently, PFedMe ~\cite{b26:t2020personalized} does not utilize the global model directly, but instead concurrently trains the global model and personalized models.

\subsection{Knowledge Distillation}

Hinton et al. have first introduced knowledge distillation ~\cite{b27:hinton2015distilling}. Their work employs a weighted sum of the hard and soft loss as the complete loss. The soft loss is the loss between the soft outputs of the student model and the soft labels generated by the teacher model, and the hard loss is the loss between the hard outputs of the student model and the real labels. Adriana Romero et al. ~\cite{b28:adriana2015fitnets} introduce knowledge distillation based on hidden layer knowledge features (hints). Zhang et al. ~\cite{b29:zhang2018deep} propose mutual distillation, enabling different models to mutually distill knowledge from one another. 

\subsection{Federated Knowledge Distillation}

Knowledge Distillation has gained increasing attention to integrating with Federated Learning ~\cite{b30:mora2022knowledge}. FedMD ~\cite{b31:li2019fedmd} makes the integration based on a shared dataset to calculate mean scores that guide the knowledge distillation of each client. Instead, FD ~\cite{b32:jeong2018communication} eliminates the need for a shared dataset and allows clients to calculate prediction scores for each label on their local dataset, and the server to calculate the global mean prediction score per label, which serves as soft labels during the local distillation. FedGKT ~\cite{b33:he2020group} combines federated learning with split learning (SL)~\cite{b34:gupta2018distributed}. FedDKC ~\cite{b35:wu2024exploring} is similar to FedGKT and can reduce the gap between knowledge distributions of the heterogeneous models. Although FedGKT and FedDKC can support resource-constrained clients, both methods require local real labels to be uploaded, which compromises client privacy. Moreover, their target is to train the small model under the guidance of large models, instead of considering how to integrate knowledge extracted from different clients to update the large model efficiently and effectively. 

\section{Methodology}

\subsection{Problem Statement}
Suppose there are $N$ clients $i \ (i=1,2,…, N)$, each of which has its private dataset with labels  $j=1, 2, …, C$. The sample size of client $i$ is $n_i$. To support the classification tasks, the key goal, defined in Equation \ref{objective}, is to minimize the loss difference between the large model updated by our method suppose constrained local resources and the conventional one suppose sufficient local resources, where $\Omega$ and $\Omega_{Conv}$ are the large model trained by our method and the conventional one, respectively, $\mathcal{L}()$ is the loss function and $D$ is the test dataset. 

\begin{equation}\label{objective}
    \mathop{\arg\min}\limits_{\Omega}F(\Omega)=\frac{\mathcal{L}(\Omega,D)-\mathcal{L}(\Omega_{Conv},D)}{\lvert D \rvert} 
\end{equation}

\subsection{Motivation}
\begin{figure}
    \centering
    \includegraphics[scale=0.45]{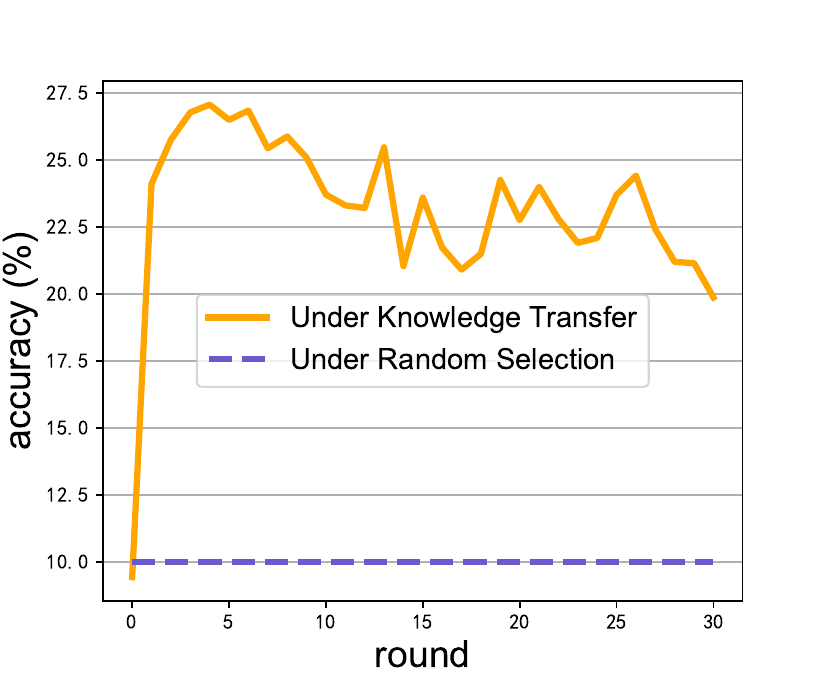}
    \caption{Accuracy ($\%$) under knowledge transfer and random selection.}
    \label{fig:motivation}
\end{figure}
Our method is based on this intuition: the small model can be viewed as the local private knowledge extractor that can be used at the server to transfer knowledge embedded within private data to the large model. 

To verify our intuition, we design a simple experiment where in each round, the small model is trained by a labeled dataset, and then a large model is fine-tuned based on a proxy dataset through knowledge distillation with the small model as the teacher model and the large model as the student model. Note that CIFAR-10 is used for small model training and its test dataset is used for evaluating the performance of the large model. Moreover, the small and large models are MobileNet V3 Small and VGG19, respectively. 

According to the result shown in Fig \ref{fig:motivation}, we can observe that the accuracy of large models can be improved significantly, even though it only processes the unlabelled proxy dataset. Therefore, the small model can share local private knowledge with the large model based on the knowledge extraction and transfer process, which motivates us to design KOALA that can integrate federated learning and knowledge distillation to implement a large-small model collaborative learning process.

\begin{figure*}
    \centering
    \includegraphics[scale=0.43]{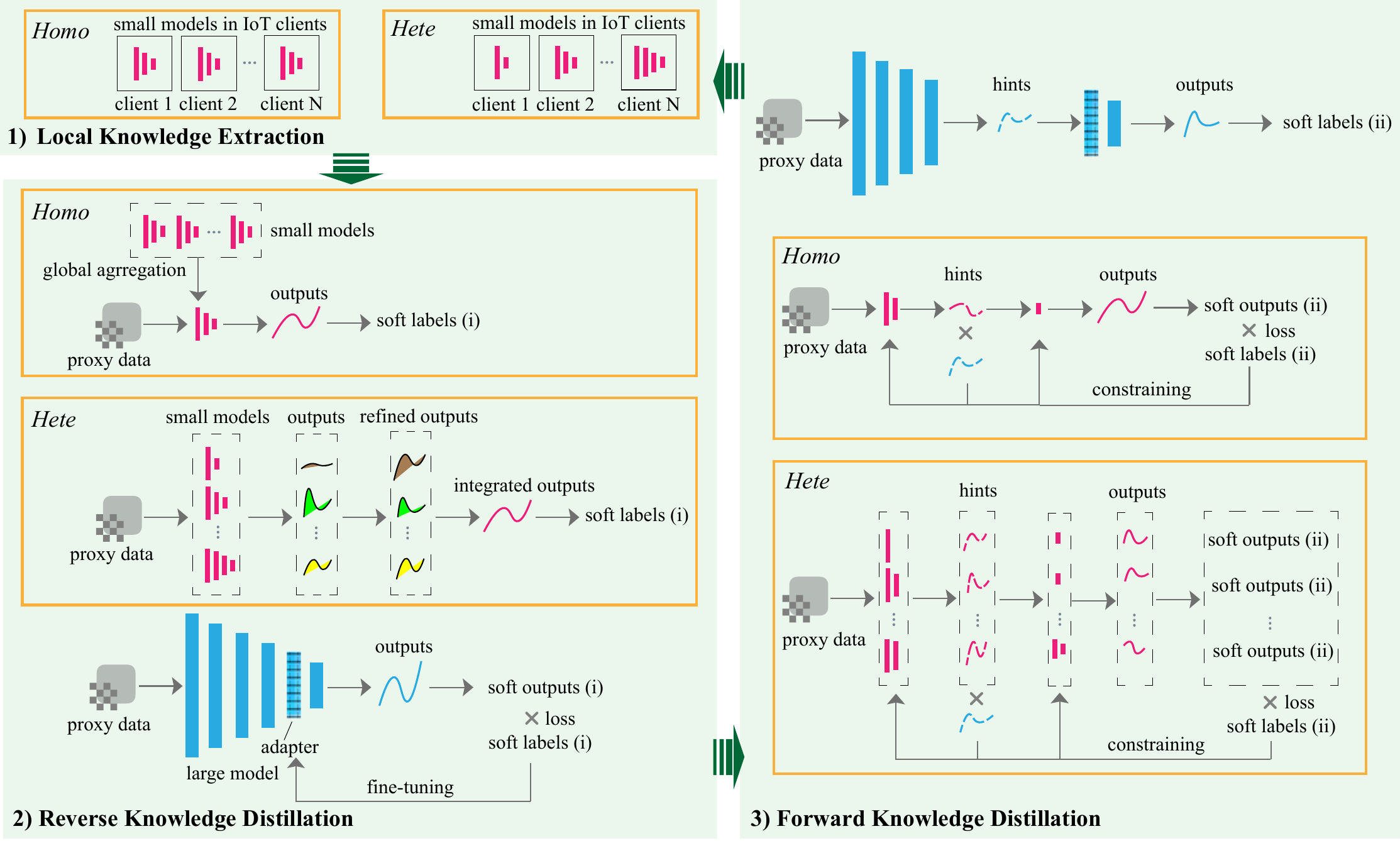}
    \caption{The framework of KOALA, which consists of 1) local knowledge extraction, 2) reverse knowledge distillation, and 3) forward knowledge distillation.}
    \label{fig:framework}
\end{figure*}

\subsection{The proposed method: KOALA}
In KOALA, we implement a large-small model collaborative learning process, through which, small models serve as local knowledge extractors and the large model is fine-tuned according to the distilled knowledge from small models. Specifically, in each IoT client, the corresponding small model is downloaded from the server and trained locally based on its private data. In the server, a bi-directional knowledge distillation mechanism is introduced, which supports 1) the reverse distillation to fine-tune the large model based on small models, and 2) the forward distillation to update small models based on the large model. 

As shown in Fig. \ref{fig:framework}, KOALA consists of three steps, namely 1) Local Knowledge Extraction, 2) Reverse Knowledge Distillation, and 3) Forward Knowledge Distillation. Since the IoT clients can be heterogeneous in not only their data but also their computing capacities, KOALA is designed with two kinds of learning modes, namely one for homogeneous small models (denoted as homo), and the other one for heterogeneous small models (denoted as hete). %%%%%%%%%%%%%%%%%%%%%%%%%%%%%%%%%%%%%%%%%%%%%%%%%%%

\subsubsection{Local Knowledge Extraction} In this step, small models either homo or hete are updated according to the private data of corresponding IoT clients. After the extraction of local knowledge, small models are uploaded to the server.                                                                                 

\subsubsection{Reverse Knowledge Distillation} 
After all the local updated small models are collected, the server starts the reverse distillation, in which, the large model serves as the student model, and the small model serves as the teacher model. 

Specifically, in the homo mode, the small models are aggregated to first generate the global small model $\omega$, and then used to produce soft labels as defined in Equation \ref{HomoSL} based on the proxy data $x$, where $T$ is distillation temperature.

\begin{equation}\label{HomoSL}
    softmax(\frac{f(x,\omega)}{T})
\end{equation}

The global small model $\omega$ transfers local knowledge to the large model $\Omega$, where the large model only updates its adapter. The reverse distillation loss $loss_{r}^{homo}$ used in the homo mode is defined in Equation \ref{HomoLoss}, where $l_{KL}()$ is KL loss function. 
\begin{equation}\label{HomoLoss}
    loss_{r}^{homo}=l_{KL}(softmax(\frac{f(x,\omega)}{T}),softmax(\frac{f(x,\Omega)}{T}))
\end{equation}

Since heterogeneous small models cannot directly be aggregated, in the hete mode, the output distributions of small models are refined and integrated to generate the consensus soft labels. 

To mediate the heterogeneity within output distributions, we introduce a distribution refinement strategy. Suppose within output distribution $f(x,\omega_{i})$, the maximum and minimum value is $z_{i,max}$, $z_{i,min}$, respectively, and the value for label $j$ is $z_{i,j}$, the refined value $\hat{z}_{i,j}$ is defined in Equation \ref{refine0}, where $\omega_{i}$ is the $i$-th small model (for client $i$), and $k$ is the coefficient to support the refinement.
\begin{equation}\label{refine0}
    \hat{z}_{i,j} = k \frac{z_{i,j}-z_{i,min}}{z_{i,max}-z_{i,min}}  
\end{equation}

To sum up the refined values for all labels, we can get
\begin{equation}\label{refine1}
    \sum_{j=1}^{C}\hat{z}_{i,j}=k \frac{\sum_{j=1}^{C}(z_{i,j}-z_{i,min})}{z_{i,max}-z_{i,min}}
    =k \frac{C(\overline{z}_{i}-z_{i,min})}{z_{i,max}-z_{i,min}}
\end{equation}

In Equation \ref{refine1}, $\overline{z}_{i}$ is the mean value of output distribution $f(x,\omega_{i})$. Suppose the mean values of refined distributions of all the small models are equal to $A$ (which is a constant), and therefore,
\begin{equation}
    A = \frac{\sum_{j=1}^{C}\hat{z}_{i,j}}{C}
    = k \frac{\overline{z}_i-z_{i,min}}{z_{i,max}-z_{i,min}}
\end{equation}

Then, the coefficient $k$ can be calculated.
\begin{equation}
    k = A \frac{z_{i,max}-z_{i,min}}{\overline{z}_i-z_{i,min}}
\end{equation}

We substitute it to Equation \ref{refine0}, and get the distribution refinement strategy as
\begin{equation}\label{refinement}
    \hat{z}_{i,j} = A \frac{z_{i,j}-z_{i,min}}{\overline{z}_{i}-z_{i,min}}
\end{equation}

According to Equation \ref{refinement}, we get the refined output distributions $\hat{z}_i=\{\hat{z}_{i,1},\hat{z}_{i,2},...,\hat{z}_{i,C}\}$. Then, we obtain the integrated output distributions among small models through Equation \ref{integration}, donated as $\widetilde{z}$. Suppose set of active clients is $S$ in this round.
\begin{equation}\label{integration}
    \widetilde{z} = \sum_{i \in S}\frac{n_i}{\sum_{i \in S}n_i}\hat{z}_{i}
\end{equation}

Based on $\widetilde{z}$, the consensus soft labels are calculated.
\begin{equation}\label{HeteSL}
    softmax(\frac{\widetilde{z}}{T})
\end{equation}

Then, we fine-tune the large model $\Omega$ based on the reverse distillation loss $loss_{r}^{hete}$ as defined in Equation \ref{HeteLoss1}.
\begin{equation}\label{HeteLoss1}
    loss_{r}^{hete} = l_{KL}(softmax(\frac{\widetilde{z}}{T}),softmax(\frac{f(x,\Omega)}{T}))
\end{equation}

\subsubsection{Forward Knowledge Distillation} 
Following the reverse distillation, we implement the forward distillation to update the small model according to the updated large model, where the large model serves as the teacher model, and the small model serves as the student model. 

To calculate the forward distillation loss, the output feature loss (the loss between the output layers) and the hidden feature loss (the loss between the hidden layers) need to be calculated. 

In the homo mode, the global small model $\omega$ is the student model to be updated. $\Omega^h$ represents the first $h$ layers within the larger model, whereas $\omega^g$ represents first $g$ layers within the global small model. Accordingly, the output feature loss $loss_{out}^{homo}$ and hidden feature loss $loss_{hid}^{homo}$ are computed according to Equations \ref{HomoOut} and \ref{HomoHid}, respectively, where $W$ is the bridging matrix and $l_{MSE}()$ is MSE loss function. 
\begin{equation}\label{HomoOut}
    loss_{out}^{homo} = l_{KL}(softmax(\frac{f(x,\Omega)}{T}),softmax(\frac{f(x,\omega)}{T}))
\end{equation}

\begin{equation}\label{HomoHid}
    loss_{hid}^{homo} = l_{MSE}(f(x,\Omega^{h}),f(x,\omega^{g})W)
\end{equation}

Therefore, the sum of $loss_{out}^{homo}$ and $loss_{hid}^{homo}$ forms the forward distillation loss $loss_{f}^{homo}$ as defined in Equation \ref{HomoLoss2}, where $\lambda$ is a constant.
\begin{equation}\label{HomoLoss2}
    loss_{f}^{homo} = loss_{out}^{homo}  + \lambda loss_{hid}^{homo} 
\end{equation}

In the hete mode, each small model $\omega_i (i\in S)$ serves as the student model undergoing knowledge distillation for the update. Suppose the $i$-th small model $\omega_i$ is the student model, $\omega^{g}_{i}$ represents first g layers within $\omega_i$ and $W_{i}$ is the bridging matrix for $\omega_i$, the output feature loss $loss_{out,i}^{hete}$ and hidden feature loss $loss_{hid,i}^{hete}$ for the $i$-th small model $\omega_i$ can be calculated according to Equations \ref{HeteOut} and \ref{HeteHid}, respectively. 

\begin{equation}\label{HeteOut}
    loss_{out,i}^{hete} = l_{KL}(softmax(\frac{f(x,\Omega)}{T}),softmax(\frac{f(x,\omega_{i})}{T}))
\end{equation}

\begin{equation}\label{HeteHid}
    loss_{hid,i}^{hete} = l_{MSE}(f(x,\Omega^{h}),f(x,\omega^{g}_{i})W_{i})
\end{equation}

Accordingly, the forward distillation loss for the $i$-th small model $loss_{f,i}^{hete}$ is 

\begin{equation}\label{HeteLoss2}
    loss_{f,i}^{hete} = loss_{out,i}^{hete} + \lambda loss_{hid,i}^{hete}
\end{equation}

Finally, either in homo or hete mode, the small model is updated based on its forward distillation loss and after the update, it is dispatched to the related client to start a new learning round until certain criteria are met (e.g., the model converges or the maximum learning round is reached).

To better illustrate the overall workflow of KOALA, its pseudo-code is given in Alg. \ref{algorithm}.

%%%%%%%%%%%%%%%%%%%%%PseudoCode%%%%%%%%%%%%%%%%%%%
\begin{algorithm}[tb]
    \caption{KOALA}
    \label{algorithm}
    \textbf{Input}: large model $\Omega$, global small model $\omega$, local small model $\omega_i$ (for client $i$), learning rate $\eta_0$,$\eta_1$,$\eta_2$, number of rounds $R$, current round $r$, set of active clients $S$, proxy data $x$, local data $(x_0,y_0)$, $l_{CE}()$ is Cross-Entropy loss function
    \begin{algorithmic}[1] %[1] enables line numbers
        \STATE Let $r=0$.
        \WHILE{$r \leq R$}
        \STATE
        $r \leftarrow r + 1$\\
        $S$ $\leftarrow$ Sampling\\
        \textbf{Client $i \in S$ executes:}\\
        $\omega_i \leftarrow \omega_i - \eta_0 
        \nabla l_{CE}(y_0,f(x_0,\omega_i))$\\
        uploading $\omega_i$ to the Server\\
        \textbf{Server executes:}\\
            
            \IF {Homo}
            \STATE
            $\omega \leftarrow \sum_{i\in S}\frac{n_i}{\sum_{i \in S}n_i}\omega_i$\\
            soft labels of $\omega$ are represented as (\ref{HomoSL})\\
            fine-tuning the large model:\\
            $loss_{r}^{homo}$ is computed as (\ref{HomoLoss})\\
            $\Omega \leftarrow \Omega - \eta_1 \nabla loss_{r}^{homo}$\\
            constraining the global small model:\\
            $loss_{f}^{homo}$ is computed as (\ref{HomoOut})(\ref{HomoHid})(\ref{HomoLoss2})\\
            $\omega \leftarrow \omega - \eta_2 \nabla loss_{f}^{homo}$
            \ENDIF

            \IF {Hete}
            \STATE
            output distributions are refined as (\ref{refinement})\\
            integrated output distributions are computed as (\ref{integration})\\
            consensus soft labels are represented as (\ref{HeteSL})\\
            fine-tuning the large model:\\
            $loss_{r}^{hete}$ is computed as (\ref{HeteLoss1})\\
            $\Omega \leftarrow \Omega - \eta_1 \nabla loss_{r}^{hete}$ \\
            constraining the small models:\\
            $loss_{f,i}^{hete}$ is computed as (\ref{HeteOut})(\ref{HeteHid})(\ref{HeteLoss2})\\
            $\omega_i \leftarrow \omega_i - \eta_2 \nabla loss_{f,i}^{hete}$
            \ENDIF
            
        \ENDWHILE
    \end{algorithmic}
\end{algorithm}

\section{Experimental Results}

%%%%%%%%%%%%%%%%%%%%%%%%%%%%%%
\begin{figure*}[!t]
\centering
\subfloat[CIFAR-10]{
		\includegraphics[scale=0.44]{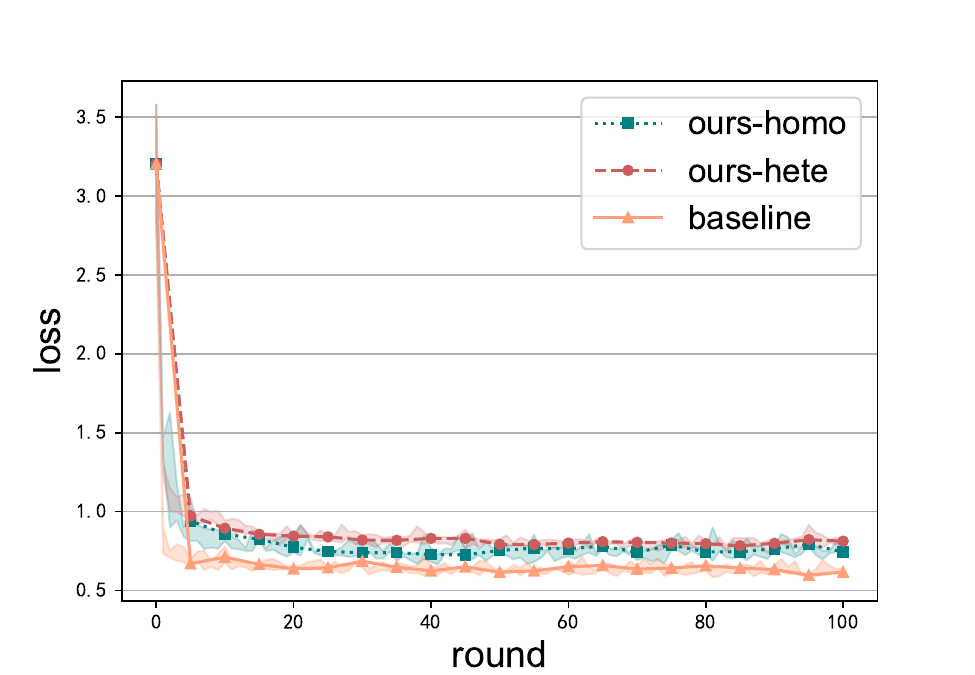}}
\subfloat[Fashion-MNIST]{
		\includegraphics[scale=0.44]{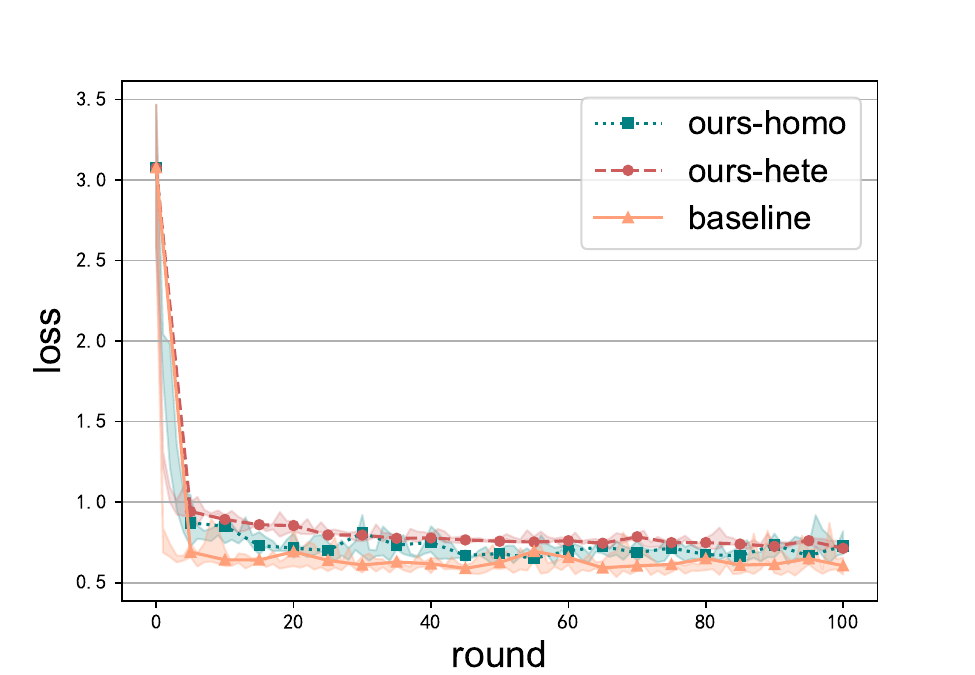}}
\\
\subfloat[USPS]{
		\includegraphics[scale=0.44]{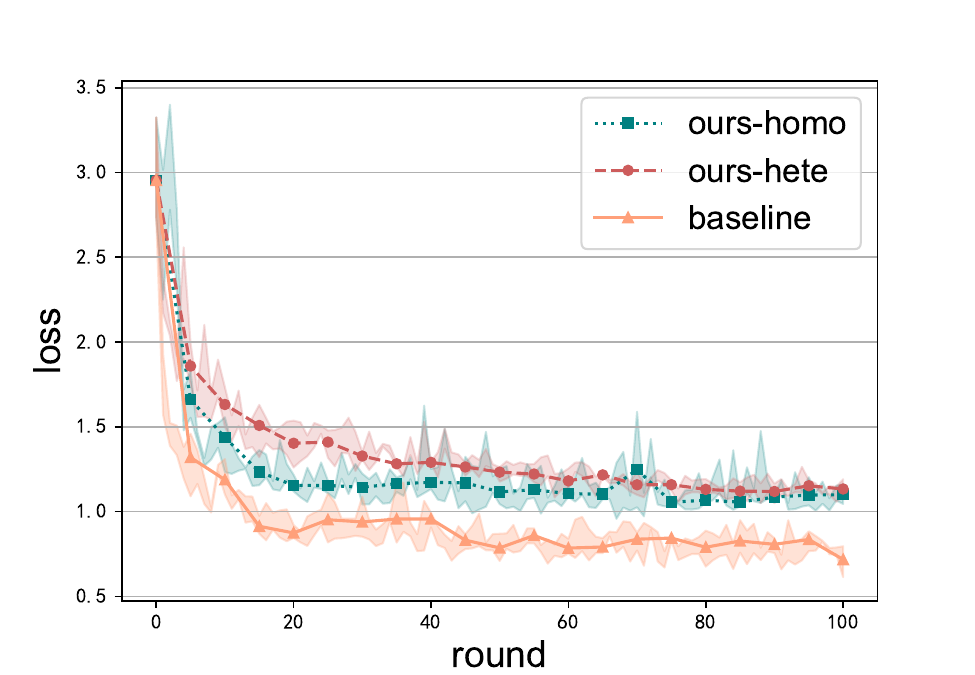}}
\subfloat[GTSRB]{
		\includegraphics[scale=0.44]{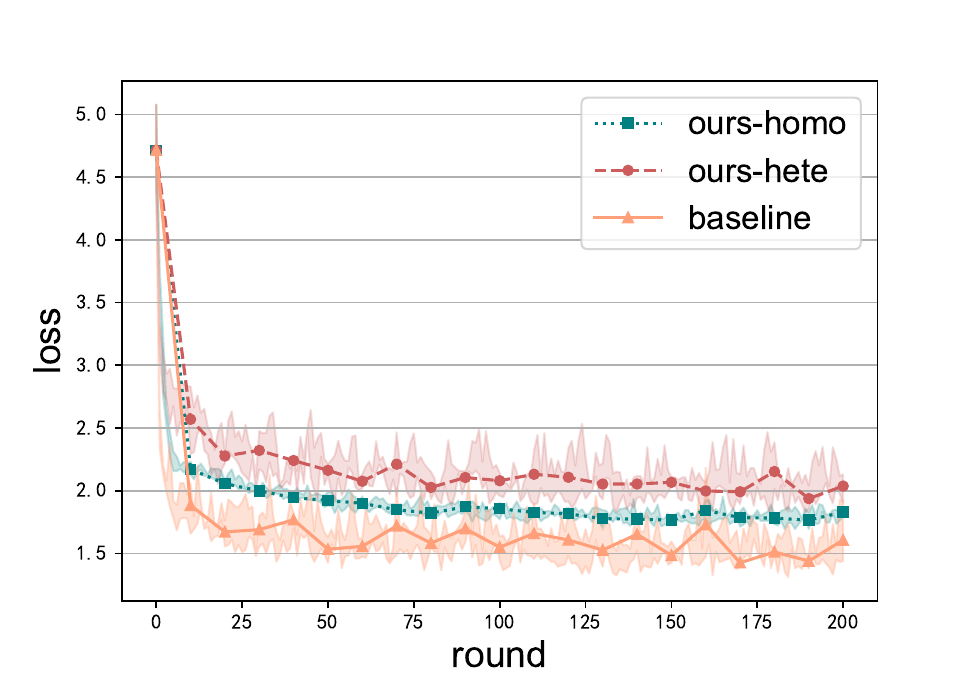}}
\caption{Loss reduction. The loss value represents the mean loss over 3 trials. The shadows indicate the range of losses across the 3 trials.}
\label{fig:loss}
\end{figure*}
%%%%%%%%%%%%%%%%%%%%%%%%%%%%%

\begin{table*}
    \centering
    \begin{tabular}{ccccc}
    \toprule
    method & CIFAR-10 & Fashion-MNIST & USPS  & GTSRB \\
    \midrule
    ours-homo & 76.02$\pm$0.55  & 77.53$\pm$0.33  & 77.76$\pm$2.01  & 51.99$\pm$1.08  \\
    ours-hete & 75.97$\pm$0.33  & 77.89$\pm$0.08  & 76.20$\pm$0.68  & 52.32$\pm$0.93  \\
    baseline & 79.35$\pm$0.12  & 80.44$\pm$0.32  & 81.48$\pm$0.88  & 58.84$\pm$0.14  \\
    \bottomrule
    \end{tabular}%
    \caption{Accuracy ($\%$). We run 3 trials and report the mean and standard derivation of the best accuracy in each trial.}
    \label{tab:acc}
\end{table*}

\begin{table*}[htbp]
  \centering
    \begin{tabular}{clllc}
    \toprule
    Model ID & \multicolumn{1}{c}{Backbone Name} & \multicolumn{1}{c}{PARAMS} & \multicolumn{1}{c}{FLOPs} & Model Type \\
    \midrule
    0     & VGG19 & 143.68M/143.71M & 3.43G/3.43G & large model \\
\cmidrule{2-5}    1     & ShuffleNet V2 X2\_0 & 7.40M/7.44M & 101.45M/101.52M & \multirow{5}[2]{*}{small model} \\
    2     & EfficientNet B0 & 5.30M/5.33M & 71.32M/71.39M &  \\
    3     & MobileNet V2 & 3.51M/3.55M & 55.84M/55.90M &  \\
    4     & MobileNet V3 Small & 2.55M/2.59M & 14.03M/14.10M &  \\
    5     & ShuffleNet V2 X0\_5 & 1.38M/1.41M & 9.18M/9.24M &  \\
    \bottomrule
    \end{tabular}%
 \caption{Model Params and FLOPs. The ID 0 and ID 1$\sim$5 respectively refers to the large model and the small models. The model of ID 3 is used for local loading and execution in ours-homo, and models of ID 1$\sim$5 are used for those in ours-hete. The large model is used for local loading and execution in baseline. In 'Params' and 'FLOPs' columns, the delimiter characters are used to separate the value of the model for CIFAR-10/Fashion-MNIST/USPS tasks (left) from that for GTSRB task (right).}
  \label{tab:PARAMSandFLOPs}%
\end{table*}%

\subsection{Setup}
We introduce the experimental setup in 4 key aspects: models, datasets, baseline, and hyperparameters.

\textbf{Models.} We select TorchVision backbones\footnote{\url{https://pytorch.org/vision/0.13/models.html}} and append the classifier onto the last layer of each backbone to form the large model and small models used in our experiments. The classifier of the large model is viewed as the adapter. The backbone for the large model is VGG19. In our homo mode, the small model is MobileNet V2. In our hete mode, the small models are MobileNet V2, MobileNet V3 Small, EfficientNet B0, ShuffleNet V2 X0\_5, and ShuffleNet V2 X2\_0, respectively. Moreover, we implement additional experiments to count the model FLOPs, where we use 64×64 randomly generated ``image'' as the input.

\textbf{Datasets.} We select 4 datasets: CIFAR-10~\cite{cifar10}, Fashion-MNIST~\cite{fmnist}, USPS~\cite{usps}, and GTSRB~\cite{gtsrb}. The entire test set of each dataset is used to evaluate the large model, recording its performance before training (round 0) and at the end of each learning round. The proxy dataset is a subset of the original train set by removing the labels. The local datasets of clients are obtained by Dirichlet Distribution, with the concentration parameter of 1.0. In addition, there is no overlap between the proxy dataset and private client datasets.

\textbf{Baseline.} We set a baseline under the assumption that all IoT clients have sufficient local resources to run the large model directly, and Federated Averaging (FedAvg) ~\cite{b17:mcmahan2017communication} is used to update the global model. Specifically, the workflow of the baseline to update the global model consists of three steps, namely: 1) clients download the global large model; 2) the large model is fine-tuned locally;  and 3) the large model parameters are uploaded to the server for the global aggregation. During client-server interactions, the adapter instead of the entire model is exchanged between the server and clients, except for the first download of the large model from the server to the clients. 

\textbf{Hyperparameters.} We consider a scenario involving 5 clients and 1 server. Adam is selected as the optimizer and the distillation temperature is set to 7 in all the experiments. For the baseline, the learning rate for local fine-tuning is 0.001, and weight decay is 0.000001. In KOALA, for reverse distillation, the learning rate is 0.001, and the weight decay is 0.000001; and for forward distillation, the learning rate  is 0.0001, and the weight decay is 0.000001. For the output distribution refinement in the hete mode, the mean value $A$ is set to 2.

\subsection{Loss and Accuracy}

The loss curves are illustrated in Figure \ref{fig:loss}, and the accuracy of different methods is listed in Table \ref{tab:acc}. It is remarkable fact that our method demonstrates optimal performance in the CIFAR-10 and Fashion-MNIST tasks, closely approaching the baseline both in loss reduction and model accuracy.

\subsection{Ablation in Bi-directional Distillation}
\begin{figure}
    \centering
    \includegraphics[scale=0.42]{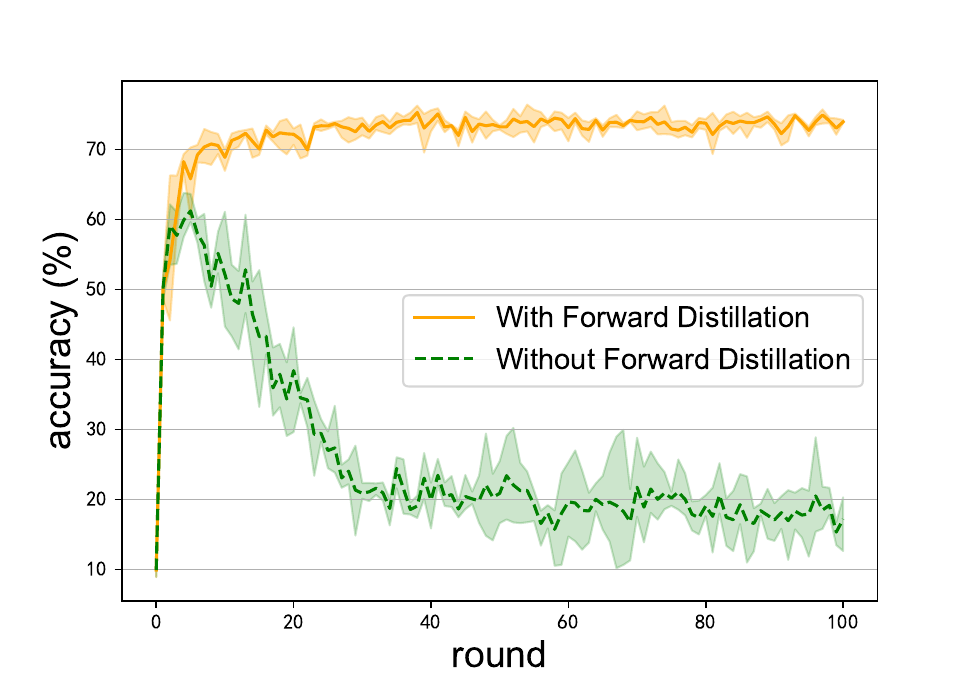}
    \caption{Accuracy ($\%$) with and without Forward Distillation. The acc is the mean value of accuracies in 3 trials and the shadows demonstrate the variability range under different random seeds.}
    \label{fig:ablation}
\end{figure}

During the bi-directional distillation in the server, the small model transfers local knowledge to the large model in reverse distillation, and the large model updates the small model in forward distillation. Reverse distillation is indispensable for fine-tuning the large model, and forward distillation also matters for the update of small models, which work jointly making the iterative learning between large and small models workable.

To reveal the necessity and efficacy of forward distillation, we implement an additional experiment with or without forward distillation in the homo mode to support the CIFAR-10 task. The experiment runs for 3 trials by using the same seed setups as the previous experiments. As illustrated in Fig \ref{fig:ablation}, it shows that forward distillation plays a significant role in the bi-directional distillation to enable the extraction of private knowledge from local clients to continuously update the large model.

\subsection{Demands for Storage and Computing Power}

\begin{figure}
    \centering
    \includegraphics[scale=0.245]{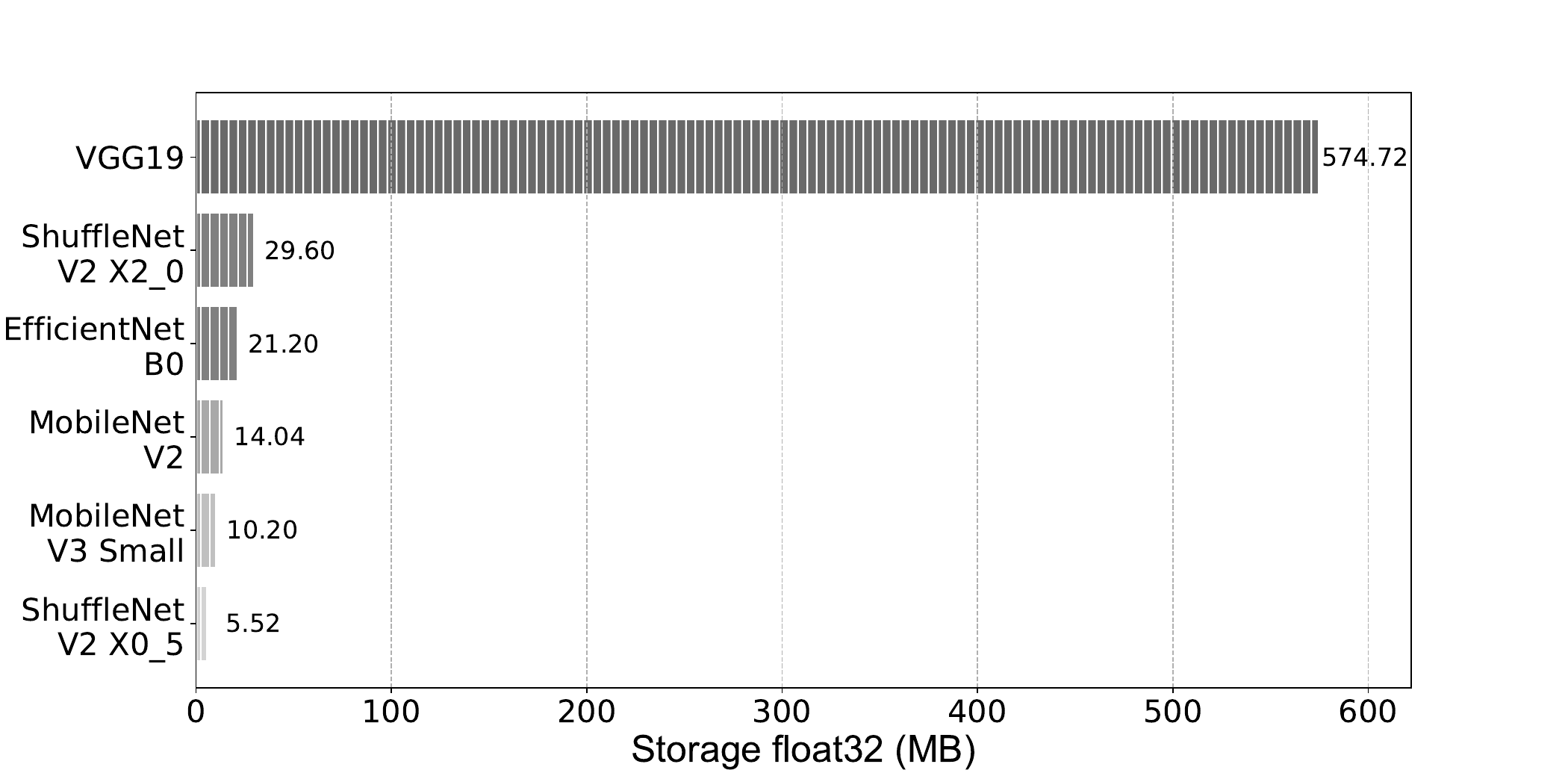}
    \caption{Storage for model local-loading in float32. The storage data in the figure is calculated according to model Params (for the CIFAR-10/Fashion-MNIST/USPS tasks) as listed in Table \ref{tab:PARAMSandFLOPs}.}
    \label{fig:loading}
\end{figure}

Table \ref{tab:PARAMSandFLOPs} shows the Params and FLOPs of the models during the experiments. The classifiers for different tasks may have a slight difference. When we demonstrate the Params and FLOPs, we use the delimiters to separate the value of the model for the CIFAR-10, Fashion-MNIST, and USPS tasks from that for the GTSRB task. In addition, Fig \ref{fig:loading} shows the storage space needed to load related models to be trained.

Since the small models have much fewer parameters than the large model, the mean storage space for all clients reduces by 97.6$\%$ (Homo) and 97.2$\%$ (Hete). We can also observe that FLOPs of the large model is significantly higher than that of each small model. The mean FLOPs of the local models of all clients (calculated according to models for CIFAR-10/Fashion-MNIST/USPS tasks) reduces by 98.4$\%$ (Homo) and 98.6$\%$ (Hete).

In summary, the storage and computing power required for the model to be loaded and executed locally are much lower than the ones needed for the baseline, which proofs the efficiency and effectiveness of KOALA in supporting various IoT scenarios consisting of large amount of heterogeneous IoT clients.

\section{Conclusion}

To fine-tune large models by orchestrating distributed IoT clients with limited storage space or computing capabilities, we propose KOALA, a privacy-preserving and resource-efficient method that integrates federated learning and knowledge distillation by implementing a novel large-small model collaborative learning process. In general, it uses small models to extract private knowledge without having large models running on IoT clients. Moreover, it also supports the knowledge transfer between the large model and small models by implementing a bi-directional distillation, in which, small models can be updated according to the large model through the common forward distillation, and also the large model can be fine-tuned by reverse distillation by aggregating knowledge from either homogeneous or heterogeneous small models. Experimental results show that compared to the conventional method, KOALA can significantly reduce the demands for local storage space and computing power to fine-tune large models with competitive performance.

\section*{Acknowledgments}
This work was supported in part by the Guangdong Basic and Applied Basic Research Foundation under Grant 2023A1515012895, in part by the National Key Research and Development Program of China under Grant 2023YFB4301900, and in part by Department of Science and Technology of Guangdong Province (Project No. 2021QN02S161).

\bibliographystyle{named}
% \bibliography{ijcai24}

\begin{thebibliography}{}

\bibitem[\protect\citeauthoryear{Adriana \bgroup \em et al.\egroup }{2015}]{b28:adriana2015fitnets}
Romero Adriana, Ballas Nicolas, K~Samira Ebrahimi, Chassang Antoine, Gatta Carlo, and Bengio Yoshua.
\newblock Fitnets: Hints for thin deep nets.
\newblock {\em Proc. ICLR}, 2(3):1, 2015.

\bibitem[\protect\citeauthoryear{Brown \bgroup \em et al.\egroup }{2020}]{b5:brown2020language}
Tom Brown, Benjamin Mann, Nick Ryder, Melanie Subbiah, Jared~D Kaplan, Prafulla Dhariwal, Arvind Neelakantan, Pranav Shyam, Girish Sastry, Amanda Askell, et~al.
\newblock Language models are few-shot learners.
\newblock {\em Advances in neural information processing systems}, 33:1877--1901, 2020.

\bibitem[\protect\citeauthoryear{Chen \bgroup \em et al.\egroup }{2023}]{b8:chen2023federated}
Chaochao Chen, Xiaohua Feng, Jun Zhou, Jianwei Yin, and Xiaolin Zheng.
\newblock Federated large language model: A position paper.
\newblock {\em arXiv preprint arXiv:2307.08925}, 2023.

\bibitem[\protect\citeauthoryear{Devlin \bgroup \em et al.\egroup }{2018}]{b1:devlin2018bert}
Jacob Devlin, Ming-Wei Chang, Kenton Lee, and Kristina Toutanova.
\newblock Bert: Pre-training of deep bidirectional transformers for language understanding.
\newblock {\em arXiv preprint arXiv:1810.04805}, 2018.

\bibitem[\protect\citeauthoryear{Dosovitskiy \bgroup \em et al.\egroup }{2020}]{b7:dosovitskiy2020image}
Alexey Dosovitskiy, Lucas Beyer, Alexander Kolesnikov, Dirk Weissenborn, Xiaohua Zhai, Thomas Unterthiner, Mostafa Dehghani, Matthias Minderer, Georg Heigold, Sylvain Gelly, et~al.
\newblock An image is worth 16x16 words: Transformers for image recognition at scale.
\newblock {\em arXiv preprint arXiv:2010.11929}, 2020.

\bibitem[\protect\citeauthoryear{Fallah \bgroup \em et al.\egroup }{2020}]{b24:fallah2020personalized}
Alireza Fallah, Aryan Mokhtari, and Asuman Ozdaglar.
\newblock Personalized federated learning: A meta-learning approach.
\newblock {\em arXiv preprint arXiv:2002.07948}, 2020.

\bibitem[\protect\citeauthoryear{Finn \bgroup \em et al.\egroup }{2017}]{b25:finn2017model}
Chelsea Finn, Pieter Abbeel, and Sergey Levine.
\newblock Model-agnostic meta-learning for fast adaptation of deep networks.
\newblock In {\em International conference on machine learning}, pages 1126--1135. PMLR, 2017.

\bibitem[\protect\citeauthoryear{Gupta and Raskar}{2018}]{b34:gupta2018distributed}
Otkrist Gupta and Ramesh Raskar.
\newblock Distributed learning of deep neural network over multiple agents.
\newblock {\em Journal of Network and Computer Applications}, 116:1--8, 2018.

\bibitem[\protect\citeauthoryear{Han \bgroup \em et al.\egroup }{2024}]{b15:han2024parameter}
Zeyu Han, Chao Gao, Jinyang Liu, Sai~Qian Zhang, et~al.
\newblock Parameter-efficient fine-tuning for large models: A comprehensive survey.
\newblock {\em arXiv preprint arXiv:2403.14608}, 2024.

\bibitem[\protect\citeauthoryear{He \bgroup \em et al.\egroup }{2020}]{b33:he2020group}
Chaoyang He, Murali Annavaram, and Salman Avestimehr.
\newblock Group knowledge transfer: Federated learning of large cnns at the edge.
\newblock {\em Advances in Neural Information Processing Systems}, 33:14068--14080, 2020.

\bibitem[\protect\citeauthoryear{Hinton \bgroup \em et al.\egroup }{2015}]{b27:hinton2015distilling}
Geoffrey Hinton, Oriol Vinyals, and Jeff Dean.
\newblock Distilling the knowledge in a neural network.
\newblock {\em arXiv preprint arXiv:1503.02531}, 2015.

\bibitem[\protect\citeauthoryear{Houlsby \bgroup \em et al.\egroup }{2019}]{b14:houlsby2019parameter}
Neil Houlsby, Andrei Giurgiu, Stanislaw Jastrzebski, Bruna Morrone, Quentin De~Laroussilhe, Andrea Gesmundo, Mona Attariyan, and Sylvain Gelly.
\newblock Parameter-efficient transfer learning for nlp.
\newblock In {\em International conference on machine learning}, pages 2790--2799. PMLR, 2019.

\bibitem[\protect\citeauthoryear{Hull}{1994}]{usps}
Jonathan~J. Hull.
\newblock A database for handwritten text recognition research.
\newblock {\em IEEE Transactions on pattern analysis and machine intelligence}, 16(5):550--554, 1994.

\bibitem[\protect\citeauthoryear{Imteaj \bgroup \em et al.\egroup }{2021}]{b11:imteaj2021survey}
Ahmed Imteaj, Urmish Thakker, Shiqiang Wang, Jian Li, and M~Hadi Amini.
\newblock A survey on federated learning for resource-constrained iot devices.
\newblock {\em IEEE Internet of Things Journal}, 9(1):1--24, 2021.

\bibitem[\protect\citeauthoryear{Jang \bgroup \em et al.\egroup }{2022}]{b23:jang2022fedclassavg}
Jaehee Jang, Heoneok Ha, Dahuin Jung, and Sungroh Yoon.
\newblock Fedclassavg: Local representation learning for personalized federated learning on heterogeneous neural networks.
\newblock In {\em Proceedings of the 51st International Conference on Parallel Processing}, pages 1--10, 2022.

\bibitem[\protect\citeauthoryear{Jeong \bgroup \em et al.\egroup }{2018}]{b32:jeong2018communication}
Eunjeong Jeong, Seungeun Oh, Hyesung Kim, Jihong Park, Mehdi Bennis, and Seong-Lyun Kim.
\newblock Communication-efficient on-device machine learning: Federated distillation and augmentation under non-iid private data.
\newblock {\em arXiv preprint arXiv:1811.11479}, 2018.

\bibitem[\protect\citeauthoryear{Karimireddy \bgroup \em et al.\egroup }{2020}]{b20:karimireddy2020scaffold}
Sai~Praneeth Karimireddy, Satyen Kale, Mehryar Mohri, Sashank Reddi, Sebastian Stich, and Ananda~Theertha Suresh.
\newblock Scaffold: Stochastic controlled averaging for federated learning.
\newblock In {\em International conference on machine learning}, pages 5132--5143. PMLR, 2020.

\bibitem[\protect\citeauthoryear{Krizhevsky \bgroup \em et al.\egroup }{2009}]{cifar10}
Alex Krizhevsky, Geoffrey Hinton, et~al.
\newblock Learning multiple layers of features from tiny images.
\newblock 2009.

\bibitem[\protect\citeauthoryear{Li and Wang}{2019}]{b31:li2019fedmd}
Daliang Li and Junpu Wang.
\newblock Fedmd: Heterogenous federated learning via model distillation.
\newblock {\em arXiv preprint arXiv:1910.03581}, 2019.

\bibitem[\protect\citeauthoryear{Li \bgroup \em et al.\egroup }{2020}]{b19:li2020federated}
Tian Li, Anit~Kumar Sahu, Manzil Zaheer, Maziar Sanjabi, Ameet Talwalkar, and Virginia Smith.
\newblock Federated optimization in heterogeneous networks.
\newblock {\em Proceedings of Machine learning and systems}, 2:429--450, 2020.

\bibitem[\protect\citeauthoryear{Li \bgroup \em et al.\egroup }{2021}]{b21:li2021model}
Qinbin Li, Bingsheng He, and Dawn Song.
\newblock Model-contrastive federated learning.
\newblock In {\em Proceedings of the IEEE/CVF conference on computer vision and pattern recognition}, pages 10713--10722, 2021.

\bibitem[\protect\citeauthoryear{Liu \bgroup \em et al.\egroup }{2019}]{b2:liu2019roberta}
Yinhan Liu, Myle Ott, Naman Goyal, Jingfei Du, Mandar Joshi, Danqi Chen, Omer Levy, Mike Lewis, Luke Zettlemoyer, and Veselin Stoyanov.
\newblock Roberta: A robustly optimized bert pretraining approach.
\newblock {\em arXiv preprint arXiv:1907.11692}, 2019.

\bibitem[\protect\citeauthoryear{McMahan \bgroup \em et al.\egroup }{2017}]{b17:mcmahan2017communication}
Brendan McMahan, Eider Moore, Daniel Ramage, Seth Hampson, and Blaise~Aguera y~Arcas.
\newblock Communication-efficient learning of deep networks from decentralized data.
\newblock In {\em Artificial intelligence and statistics}, pages 1273--1282. PMLR, 2017.

\bibitem[\protect\citeauthoryear{Mora \bgroup \em et al.\egroup }{2022a}]{b18:mora2022federated}
Alessio Mora, Davide Fantini, and Paolo Bellavista.
\newblock Federated learning algorithms with heterogeneous data distributions: An empirical evaluation.
\newblock In {\em 2022 IEEE/ACM 7th Symposium on Edge Computing (SEC)}, pages 336--341. IEEE, 2022.

\bibitem[\protect\citeauthoryear{Mora \bgroup \em et al.\egroup }{2022b}]{b30:mora2022knowledge}
Alessio Mora, Irene Tenison, Paolo Bellavista, and Irina Rish.
\newblock Knowledge distillation for federated learning: a practical guide.
\newblock {\em arXiv preprint arXiv:2211.04742}, 2022.

\bibitem[\protect\citeauthoryear{Radford \bgroup \em et al.\egroup }{2018}]{b3:radford2018improving}
Alec Radford, Karthik Narasimhan, Tim Salimans, Ilya Sutskever, et~al.
\newblock Improving language understanding by generative pre-training.
\newblock 2018.

\bibitem[\protect\citeauthoryear{Radford \bgroup \em et al.\egroup }{2019}]{b4:radford2019language}
Alec Radford, Jeffrey Wu, Rewon Child, David Luan, Dario Amodei, Ilya Sutskever, et~al.
\newblock Language models are unsupervised multitask learners.
\newblock {\em OpenAI blog}, 1(8):9, 2019.

\bibitem[\protect\citeauthoryear{Simonyan and Zisserman}{2014}]{b6:simonyan2014very}
Karen Simonyan and Andrew Zisserman.
\newblock Very deep convolutional networks for large-scale image recognition.
\newblock {\em arXiv preprint arXiv:1409.1556}, 2014.

\bibitem[\protect\citeauthoryear{Stallkamp \bgroup \em et al.\egroup }{2012}]{gtsrb}
Johannes Stallkamp, Marc Schlipsing, Jan Salmen, and Christian Igel.
\newblock Man vs. computer: Benchmarking machine learning algorithms for traffic sign recognition.
\newblock {\em Neural networks}, 32:323--332, 2012.

\bibitem[\protect\citeauthoryear{T~Dinh \bgroup \em et al.\egroup }{2020}]{b26:t2020personalized}
Canh T~Dinh, Nguyen Tran, and Josh Nguyen.
\newblock Personalized federated learning with moreau envelopes.
\newblock {\em Advances in Neural Information Processing Systems}, 33:21394--21405, 2020.

\bibitem[\protect\citeauthoryear{Tan \bgroup \em et al.\egroup }{2022}]{b22:tan2022towards}
Alysa~Ziying Tan, Han Yu, Lizhen Cui, and Qiang Yang.
\newblock Towards personalized federated learning.
\newblock {\em IEEE Transactions on Neural Networks and Learning Systems}, 2022.

\bibitem[\protect\citeauthoryear{Wang \bgroup \em et al.\egroup }{2019}]{b10:wang2019edge}
Xiaofei Wang, Yiwen Han, Chenyang Wang, Qiyang Zhao, Xu~Chen, and Min Chen.
\newblock In-edge ai: Intelligentizing mobile edge computing, caching and communication by federated learning.
\newblock {\em Ieee Network}, 33(5):156--165, 2019.

\bibitem[\protect\citeauthoryear{Wu \bgroup \em et al.\egroup }{2022}]{b13:wu2022communication}
Chuhan Wu, Fangzhao Wu, Lingjuan Lyu, Yongfeng Huang, and Xing Xie.
\newblock Communication-efficient federated learning via knowledge distillation.
\newblock {\em Nature communications}, 13(1):2032, 2022.

\bibitem[\protect\citeauthoryear{Wu \bgroup \em et al.\egroup }{2024}]{b35:wu2024exploring}
Zhiyuan Wu, Sheng Sun, Yuwei Wang, Min Liu, Quyang Pan, Junbo Zhang, Zeju Li, and Qingxiang Liu.
\newblock Exploring the distributed knowledge congruence in proxy-data-free federated distillation.
\newblock {\em ACM Transactions on Intelligent Systems and Technology}, 15(2):1--34, 2024.

\bibitem[\protect\citeauthoryear{Xiao \bgroup \em et al.\egroup }{2017}]{fmnist}
Han Xiao, Kashif Rasul, and Roland Vollgraf.
\newblock Fashion-mnist: a novel image dataset for benchmarking machine learning algorithms.
\newblock {\em arXiv preprint arXiv:1708.07747}, 2017.

\bibitem[\protect\citeauthoryear{Yu \bgroup \em et al.\egroup }{2023}]{b12:yu2023federated}
Sixing Yu, J~Pablo Mu{\~n}oz, and Ali Jannesari.
\newblock Federated foundation models: Privacy-preserving and collaborative learning for large models.
\newblock {\em arXiv preprint arXiv:2305.11414}, 2023.

\bibitem[\protect\citeauthoryear{Zhang \bgroup \em et al.\egroup }{2018}]{b29:zhang2018deep}
Ying Zhang, Tao Xiang, Timothy~M Hospedales, and Huchuan Lu.
\newblock Deep mutual learning.
\newblock In {\em Proceedings of the IEEE conference on computer vision and pattern recognition}, pages 4320--4328, 2018.

\bibitem[\protect\citeauthoryear{Zhang \bgroup \em et al.\egroup }{2021}]{b16:zhang2021survey}
Chen Zhang, Yu~Xie, Hang Bai, Bin Yu, Weihong Li, and Yuan Gao.
\newblock A survey on federated learning.
\newblock {\em Knowledge-Based Systems}, 216:106775, 2021.

\bibitem[\protect\citeauthoryear{Zhuang \bgroup \em et al.\egroup }{2023}]{b9:zhuang2023foundation}
Weiming Zhuang, Chen Chen, and Lingjuan Lyu.
\newblock When foundation model meets federated learning: Motivations, challenges, and future directions.
\newblock {\em arXiv preprint arXiv:2306.15546}, 2023.

\end{thebibliography}

\end{document}